\documentclass{article}
\usepackage{spconf,amsmath,graphicx}
\usepackage{multirow}
\usepackage{makecell}

\usepackage{booktabs}

\title{MEAT: Median-Ensemble Adversarial Training for Improving Robustness and Generalization}
\name{Zhaozhe Hu\textsuperscript{1,2}, Jia-Li Yin\textsuperscript{1,2$\ast$}\thanks{$^{\ast}$Corresponding author. Email: jlyin@fzu.edu.cn}, Bin Chen\textsuperscript{1,2}, Luojun Lin\textsuperscript{2}, Bo-Hao Chen\textsuperscript{3}, Ximeng Liu\textsuperscript{1,2$\ast$}}
\address{Fujian Province Key Laboratory of Information Security and Network Systems, Fuzhou 350108, China\textsuperscript{1}\\
College of Computer Science and Big Data, Fuzhou University, Fuzhou 350108, China\textsuperscript{2}\\
Department of Computer Science and Engineering, Yuan Ze University, Taiwan\textsuperscript{3}}

\begin{document}
%
\maketitle
\begin{abstract}
Self-ensemble adversarial training methods improve model robustness by ensembling models at different training epochs, such as model weight averaging (WA). However, previous research has shown that self-ensemble defense methods in adversarial training (AT) still suffer from robust overfitting, which severely affects the generalization performance. Empirically, in the late phases of training, the AT becomes more overfitting to the extent that the individuals for weight averaging also suffer from overfitting and produce anomalous weight values, which causes the self-ensemble model to continue to undergo robust overfitting due to the failure in removing the weight anomalies. To solve this problem, we aim to tackle the influence of outliers in the weight space in this work and propose an easy-to-operate and effective \textit{Median-Ensemble Adversarial Training} (MEAT) method to solve the robust overfitting phenomenon existing in self-ensemble defense from the source by searching for the median of the historical model weights. Experimental results show that MEAT achieves the best robustness against the powerful AutoAttack and can effectively allievate the robust overfitting. We further demonstrate that most defense methods can improve robust generalization and robustness by combining with MEAT.
\end{abstract}
\section{Introduction}
\label{sec:intro}

Because deep neural networks (DNNs) can be attacked by imperceptible perturbations, a number of methods~\cite{madry2017pgd,tramer2017ensembleat,Yin2023} have been proposed against the adversarial attacks~\cite{szegedy2013begin_ae,croce2020autoattack}. Among all these methods, adversarial training (AT)~\cite{madry2017pgd} is widely recognized as the most effective way to improve model robustness. Unfortunately, Rice et al.~\cite{rice2020overfitting} observed that AT is suffering from robust overfitting, which is virtually nonexistent in standard training: shortly after the first learning rate decay in AT, the robust accuracy on the test set will continue to degrade as training proceeds. The early stopping method is commonly used in AT to mitigate overfitting which is simple to implement and effective. However, early stopping cannot simultaneously balance the prevention of overfitting and robustness improvement, \textit{i.e.}, stopping too early leads to the model failing to maximize its generalization capability during the training process. Therefore, for the improvement of AT, preventing robust overfitting and improving robustness are crucial.

\begin{figure}
    \centering
    \includegraphics[width=0.45\textwidth]{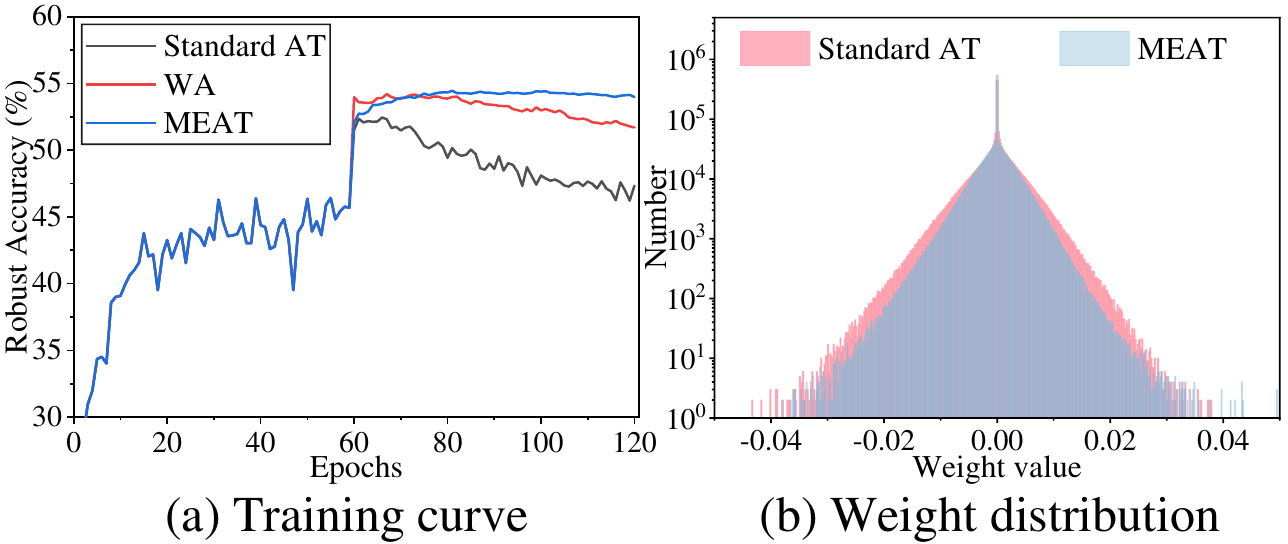}
    \caption{The training curve (a) of robust accuracy (\%) against PGD-20 attack using standard AT w./w.o. WA and MEAT and the distribution of weight values (b) in the last convolutional layer. The learning rate drops at 60 epochs. Compared to standard AT and WA, MEAT effectively mitigates robust overfitting while maintaining high robust accuracy.}
    \label{different_ensemble}
    \vspace{-10pt}
\end{figure}

Traditional ensemble defense methods~\cite{tramer2017ensembleat} have shown improved robustness by ensembling a number of models; however, adversarially training each surrogate model can lead to a huge computational burden. In contrast, Chen et al.~\cite{chen2021use_wa_overfit} proposed the self-ensemble defense method where they applied weight averaging~\cite{izmailov2018swa} (WA) on models at different training epochs to achieve a more stable model. Such self-ensemble can significantly reduce the computational burden since the models for the ensemble are the historical model during training. However, we find that although it can maintain clean accuracy, it still suffers from robust overfitting, as shown in Fig.\,\ref{different_ensemble}. This is because that when the trained model goes into the overfitting, the model weights are anomalous and using weight averaging cannot remove the anomalies, the averaged model would still suffer from the overfitting. 


In this paper, we aim to develop an ensemble method capable of preventing robust overfitting alone rather than depending on other augmentations, making it portable to combine with other state-of-the-art methods to eliminate overfitting and improve robustness. We first analyze WA in AT and find that using WA cannot well eliminate the effect of the anomalous model parameters due to the averaging operation. This phenomenon reminds us the removing of impulse noises (i.e., certain values in a signal are extremely changed) where the average filters cannot well remove such noises but median filters can eliminate the effect of outliers and work better in this situation. Inspired by this, we propose a Median-Ensemble Adversarial Training (MEAT) that optimizes the weight states of overfitting individuals by calculating the median of historical model weights to retain valid weight values and drop abnormal weight values. In such a way, the influence of the anomalous model parameters can be minimized during ensemble. We conduct extensive experiments on various datasets and find that the proposed MEAT can work much better than the WA self-ensemble. In summary, our contributions are as follows:


\begin{itemize}
    \item We analyze why WA fails in preventing robust overfitting in AT and further propose a Median-Ensemble Adversarial Training (MEAT) method which optimizes the weight states of overfitting individuals by calculating the median of historical model weights.
    \item We demonstrate the ability of MEAT to mitigate robust overfitting as well as visualize the variation in 3D loss landscapes for standard AT and MEAT, which enables the observation of generalization gap visually and strongly justifies the effectiveness of MEAT.
    \item We compare the robustness and generalization performance of MEAT and other SOTA defense methods on different datasets. The results show that MEAT achieves the most advanced accuracy and robustness, and effectively minimize the generalization gap.
\end{itemize}

\section{Preliminaries: Adversarial Training}
\label{sec:preliminaries}

The central idea of adversarial training (AT) is to make the classifier robust by directly applying adversarial examples for training~\cite{goodfellow2014fgsm,madry2017pgd}. In general, AT methods consider the training process of the classifier $f_\theta$ as the solution of a min-max problem: internally perturbing a given input to $x+\delta$ to maximize a certain classification loss $\ell$; externally optimizing the parameters $\theta$ of the classifier by minimizing the classification loss $\ell$. Here, we define the min-max process of AT as:
\vspace{-10pt}
\begin{equation}			
\min_{\theta} \mathrm{E}_{(x,y) \sim D} \left[\max_{||\delta||_{p} \leq \epsilon} \ell(f_{\theta}(x + \delta), y) \right], \label{minmax}
\end{equation}
where $(x,y)$ denotes the example sampled from the training data distribution $D$, $f_{\theta}$ denotes the model with parameters $\theta$, $\ell$ denotes the loss function, ${||\delta||_{p} \leq \epsilon}$ denotes ${l_p}$ norm-bounded perturbations $\delta$ of size $\epsilon$. In this paper, we primarily use PGD with the cross-entropy loss ${\ell}_{ce}$ to train classification $f_\theta$.

\section{Proposed Method}
\label{sec:method}

\subsection{Analysis of Model Weight Averaging}
In order to mitigate robust overfitting phenomenon and improve the robust generalization of the model on test sets, common approaches such as data augmentation~\cite{li2023daalone,rebuffi2021data}, regularization techniques~\cite{tack2022ccg,wu2020awp}, and model ensemble~\cite{wang2022seat,chen2021use_wa_overfit} have been proposed. Among them, Wu et al.~\cite{wu2020awp} showed that robust overfitting is correlated with the adversarial loss landscape~\cite{li2018losslandscape}, \textit{i.e.}, the flatter is the loss landscape, the smaller is the robust generalization gap. WA uses a fixed decay rate $\tau$ to obtain $\theta'$ by weighted average of the model parameters $\theta$ at each training step ( \textit{i.e.}, $\theta' \leftarrow \tau \cdot \theta' +(1-\tau) \cdot \theta$). Although applying WA to AT to smooth the weights and find flatter adversarial loss landscapes is sensible, Rebuffi et al.~\cite{rebuffi2021data} note that WA, while improving model robustness, still tends to robust overfitting. Unlike standard training, in the middle and late phases of AT, the models in AT all tend to overfit, which can lead to the ensemble model consisting of overfitting individuals still suffering from robust overfitting. Similarly, the deterioration has been found in the self-ensemble method proposed by Wang et al~\cite{wang2022seat}. Mitigating the robust overfitting of WA using more training data~\cite{schmidt2018moredata} is in essence reducing the number of overfitting individuals, but training costs also increase. As shown in Fig.\,\ref{different_ensemble} (a), we trained PRN-18 on CIFAR-10 and started MEAT and WA at 60 epochs. Standard AT combined with WA purely improves the best robustness of the model not alleviating robust overfitting, and the robust accuracy decreases at a speed equivalent to that of standard AT. 


\vspace{-10pt}
\subsection{Proposed Method}
Because robust overfitting in AT causes the model to overfit the known adversarial samples, which increases the number of anomalous weights in overfitting models, at this point weight averaging is no longer a solution to the problem of anomalous weights. In the late phases of training, if the anomalous weights can be dropped and the effective weights retained, the overfitting individuals can be effectively exploited thereby minimizing the robust overfitting. Specifically, overfitting leads to an increased likelihood of weight anomalies in the model (weight values that are too big or small in relation terms), and weight averaging reduces the impact of anomalies by averaging with normal weights instead of dropping the anomalous ones, while median ensembles can drop anomalies by computing the weight median because the median value is a relatively stable statistic and will only use the existing weights. Inspired by this, We introduce median instead of average for the first time to model ensemble methods and devise a median-ensemble AT method that produces a robust self-ensemble model by using the median of the weights from the historical models. The histogram of weight values is shown in Fig.\,\ref{different_ensemble} (b), where the distribution of MEAT is more compact at the last checkpoint, which indicates that many abnormal values of weights are replaced by normal values, leading to a better weight ensemble.

Formally, for given $n$ historical model weights, every model has the same architecture and contains the same number of weights at different time steps during AT. For the $k$-th layer of weights, this can be expressed as a matrix:
\vspace{-10pt}
\begin{equation}
    w_k = \begin{bmatrix} w_{1,k}, w_{2,k}, \dots, w_{n,k} \end{bmatrix},
\end{equation}
where $w_k$ is the weight of the $n$ models at the $k$-th layer. The new weight matrix obtained through median ensemble can be expressed as:
\vspace{-5pt}
\begin{equation}
    \tilde{w}_k = \mathbf{Median}(w_k),
\end{equation}
where $\tilde{w}_k$ is the new weight of the $k$-th layer and $\mathbf{Median(\cdot)}$ denotes the function that finds the median of the vector $w_k$ at the same location. This algorithm operates for every layer to obtain a new weight matrix where each element is obtained by median solving. Taken together, the new weights obtained by calculating the median can be represented as:
\vspace{-10pt}
\begin{equation}
    \tilde{w} = \begin{bmatrix} \tilde{w}_1, \tilde{w}_2, \dots, \tilde{w}_k \end{bmatrix},
\end{equation}
where $\tilde{w}$ denotes the weights of the ensemble model generated by our approach, and $k$ denotes the number of layers.
Because AT does not exhibit overfitting in the early phases, MEAT is applied to the middle and late phases of training. Notably, the number $n$ of historical models selected is gradually increased rather than fixed, which is advantageous not only to make use of the latest optimized weights but also to take into account some of the effects of the historical weights. If the DNN uses BatchNorm~\cite{ioffe2015batchnorm}, we run an additional pass over the data similar to WA and compute the running mean and standard deviation of the activations at each layer of the network with $\tilde{w}$ weights after training. 

\begin{figure}[t]
    \centering
    \includegraphics[width=0.45\textwidth]{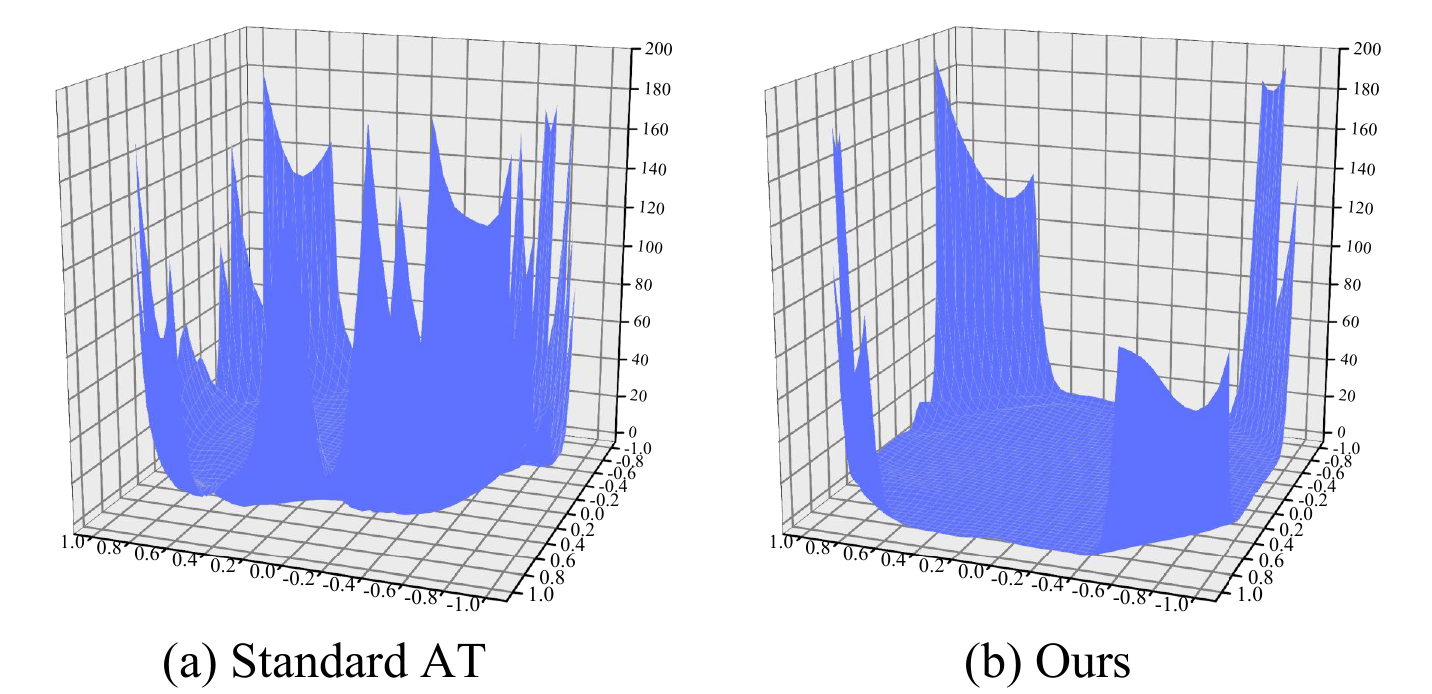}
    \caption{Comparison of the adversarial loss landscape of models trained by standard AT (a) and MEAT (b) using WRN-34-10 on CIFAR-10. $z$ axis denotes the loss value. We plot the loss landscape function: $z=\ell(\theta+\frac{v_1}{\lVert v_1 \rVert}\lVert \theta \rVert + \frac{v_2}{\lVert v_2 \rVert}\lVert \theta \rVert)$, where $v_1$ and $v_2$ denote two random vectors sampled from a Gaussian distribution, and $\lVert \cdot \rVert$ denotes the Frobenius norm.}
    \label{losslandscape}
    \vspace{-10pt}
\end{figure}

We follow the robust generalization claim of Wu et al.~\cite{wu2020awp} and delve into the evidence of that median-ensemble technique improves the loss landscape by adopting the scheme provided by Li et al.~\cite{li2018losslandscape} as shown in Fig.\,\ref{losslandscape}. The loss landscape of the standard AT model varies drastically, which means that it has higher test errors along certain directions; the loss landscape of our model is smooth and varies slowly, which improves the robust generalization ability.

\begin{table*}[t]
    \centering
    \fontsize{9}{11}\selectfont
    \caption{Comparison of clean accuracy and robust accuracy (\%) of our method with different defense methods using WRN-34-10 on CIFAR-10 and CIFAR-100. The values in parentheses denote the test accuracy of the last checkpoint.}
    
    \begin{tabular}{lccc|ccc}
    
       \toprule
       \multirow{2}{*}{Method}
       &\multicolumn{3}{c|}{CIFAR-10 dataset}
       &\multicolumn{3}{c}{CIFAR-100 dataset}\\
       \cmidrule(l){2-7}
            &Clean &PGD-20 &AutoAttack &Clean &PGD-20 &AutoAttack \\ 
             \midrule

              
              Standard AT \cite{madry2017pgd}       &85.23 (80.50) &48.93 (44.42)  &45.62 (41.99) &60.29 (53.59)  &26.84 (20.25) &22.48 (16.98)\\
              TRADES \cite{zhang2019trades}         &83.94 (82.08) &55.48 (53.19)  &51.28 (49.12) &59.52 (56.05)  &30.43 (28.78) &24.56 (23.24)\\
              MART \cite{wang2020mart}              &83.06 (84.66) &56.24 (51.78)  &53.02 (45.74) &61.72 (56.70)  &29.94 (28.13) &25.60 (24.94)\\
              AWP \cite{wu2020awp}                  &86.37 (85.90) &58.18 (56.55)  &53.51 (52.21) &63.59 (62.50)  &34.47 (34.27) &29.38 (29.07)\\
              CCG \cite{tack2022ccg}                &86.56 (84.85) &58.28 (54.13)  &52.16 (49.51) &60.74 (58.19)  &27.41 (27.27) &23.22 (22.75)\\
              IDBH \cite{li2023daalone}             &87.04 (86.20) &57.60 (56.98)  &53.04 (52.44) &60.66 (61.54)  &32.61 (31.19) &27.39 (26.42)\\
              GAIRAT \cite{zhang2020gairat}         &85.84 (85.50) &57.63 (54.50)  &43.49 (43.18) &53.61 (60.01)  &26.28 (25.63) &22.37 (21.62)\\
              SEAT \cite{wang2022seat}              &86.02 (87.55) &59.25 (51.95)  &54.82 (47.95) &62.25 (62.35)  &34.17 (26.62) &29.70 (22.08)\\
              Standard AT (with WA) \cite{madry2017pgd} &86.57 (84.21) &57.78 (54.93)  &52.79 (49.88) &63.84 (57.78)  &32.41 (29.84) &28.83 (25.19)\\
              IDBH (with WA) \cite{li2023daalone}   &88.23 (88.23) &60.92 (60.88)  &55.72 (55.70) &64.39 (65.66)  &35.70 (35.28) &30.89 (31.01)\\
              \midrule
              MEAT (ours)                           &87.38 (88.50) &60.08 (58.89)  &55.20 (54.26) &64.21 (63.87)  &35.03 (33.11) &30.64 (27.42)\\
              \textbf{MEAT+IDBH (ours)}             &\textbf{89.06} (\textbf{88.57})   &\textbf{61.08} (\textbf{60.93})  &\textbf{56.13} (\textbf{55.89})  &\textbf{64.93} (\textbf{66.07})   &\textbf{35.93} (\textbf{35.13})  &\textbf{31.12} (\textbf{31.23})\\
              \bottomrule
              
    \end{tabular}
    \label{white-box-cifar10-result}
\vspace{-8pt}
\end{table*}

\section{Experiments}
\label{sec:experiments}

\subsection{Experimental Setups}
We used CIFAR-10 and CIFAR-100~\cite{krizhevsky2009cifar} as datasets and WRN-34-10~\cite{zagoruyko2016wideresnet} as architecture. We trained the model using SGD with 0.9 momentum for 120 epochs, and the weight decay factor is set as $5e^{-4}$. The learning rate starts at 0.1, reduces linearly from one-third to two-thirds of the total epochs at 0.01, and from two-thirds to the last epochs at 0.001. We set the maximum perturbation magnitude at each pixel as $\epsilon=8/255$, step size as $2/255$, the number of steps as 10 and use the $l_\infty$ threat model. For the proposed method, we start the computation from half of the total training epochs (60 epochs) until the completion of the training. For the evaluation of different defense methods, we used PGD~\cite{madry2017pgd} and AutoAttack (AA)~\cite{croce2020autoattack} attacks. The defense methods compared all follow the official implementation for better robustness.

\vspace{-10pt}
\subsection{Performance Evaluation}

\noindent\textbf{Main Results.}
We first verify the effectiveness of the proposed method from both robustness and robust generalization perspectives. The results on CIFAR-10 and CIFAR-100 are shown in Table\,\ref{white-box-cifar10-result}. It can be clearly seen that MEAT achieves the best clean accuracy and robustness. On CIFAR-10, TRADES and MART focus on robustness improvement, resulting in a decrease in clean accuracy. In contrast, MEAT provides a good trade-off between clean and robust accuracy, even using only the traditional cross-entropy loss function for training the model. AWP and CCG reduce the robust generalization gap by the addition of regularization terms. Compared to CCG, MEAT has a smaller robust generalization gap and stable robustness in the last checkpoint. Notably, IDBH is popularized in AT by setting the hardness range of data augmentation to mitigate the robust overfitting phenomenon well. The combination with IDBH solves the robust overfitting confronted by WA while further improving robustness. However, the robustness of the model will be better than WA when combined with MEAT, which makes sense because MEAT allows for a better weight ensemble. GAIRAT mitigates robust overfitting by reallocating weights to the losses of the adversarial examples and is robust under the PGD-20 attack, but it is not effective against all attacks, failing to perform well against the ensemble attack AA. MEAT, meanwhile, solves this problem so well that it is always able to retain the best robustness under both PGD-20 and AA attacks, which shows that the robustness improvement of MEAT is comprehensive. Further, we compared standard AT combined with WA and SEAT as the single model ensemble defense with MEAT and found that their improvement for robustness is substantial but still suffers from robust overfitting, in line with our previous judgment. With regard to SEAT, the best robust accuracy is rivaled by IDBH combined with WA, but the 9\% gap in final checkpoint accuracy is fatal. It is worth noting that MEAT has obvious advantages both in terms of robustness improvement and robust generalization gap reduction. 
Our method still has the best performance on CIFAR-100, and MEAT significantly minimizes the seriousness of robust overfitting and improves the clean accuracy and robustness of standard AT. Compared to the standard AT, the best and last robustness of MEAT is improved by 8.19\% and 12.86\%, respectively.

\vspace{3pt}
\noindent\textbf{Self-ensemble Defenses Based on Weight Averaging.}
The results in Table\,\ref{white-box-cifar10-result} also show that standard AT with WA and SEAT methods both clearly display signs of robust overfitting. More specifically, we found that the degree of overfitting in SEAT, \textit{i.e.}, the gap between the robust accuracy of the last and best checkpoints, is 7.3\%, which is much larger than that in standard AT with WA of 2.85\%. This is consistent with our analysis that SEAT uses an exponential moving average (EMA) to assign weights to the historical models, which means that SEAT puts a greater weight on the most recent historical model; whereas WA uses weighted averaging, where each historical model is assigned equal weights. Such behavior leads to the fact that in the late phase of training and SEAT assigns a greater weight to individuals who suffer from overfitting compared to WA.

\begin{table}[t]
\centering
\fontsize{9}{11}\selectfont
\caption{The gap ($\Delta$) between accuracy (\%) of the best and last checkpoints and test robustness (\%) on CIFAR-10 using WRN-34-10. The values in parenthesis denote the result of the last checkpoint with the PGD-20 accuracy.}
\vspace{8pt}
\begin{tabular}{l|ccc}
\toprule
Method                                         & Clean & PGD-20 & Gap ($\Delta$)     \\
\midrule 
TRADES \cite{zhang2019trades}                  & 83.94 (82.08) & 55.48 (53.19) & 2.29    \\
\textbf{+MEAT}                                 & \textbf{86.53} (\textbf{86.82}) & \textbf{59.24} (\textbf{59.11}) & \textbf{0.13} \\
\midrule
MART \cite{wang2020mart}                       & \textbf{83.06} (84.66) & 56.24 (51.78) & 4.46    \\
\textbf{+MEAT}                                 & 79.54 (\textbf{85.12}) & \textbf{58.75} (\textbf{55.78}) & \textbf{2.97}   \\
\midrule
FAT \cite{zhang2020fat}                        & 89.78 (88.96) & 48.73 (46.58) & 2.15  \\
\textbf{+MEAT}                                 & \textbf{90.34} (\textbf{89.48}) & \textbf{52.62} (\textbf{50.99}) & \textbf{1.63} \\
\bottomrule
\end{tabular}
\label{with-our-con-white-box-cifar10-result}

\end{table}
\vspace{-12pt}
\subsection{Ablation Study}
To validate the effectiveness of our method on mitigating overfitting, we combine several defense methods with MEAT for AT, as shown in Table\,\ref{with-our-con-white-box-cifar10-result}. Overall, MEAT effectively minimizes the gap between the robust accuracy of the last and best checkpoints and considerably improves robustness under the PGD-20 attack. MART+MEAT performs less well in clean accuracy, and we speculate that the lack of good balance between the clean accuracy and robustness of MART leads the historical model to focus excessively on adversarial examples, which degrades the clean accuracy of individuals.
\vspace{-12pt}
\section{Conclusion}
\label{sec:conclusion}
\vspace{-10pt}
In this paper, we first point out the reasons for the robust overfitting of the model weight averaging method, then propose a new ensemble method called Median-Ensemble Adversarial Training (MEAT) for finding the optimal solution in the weight space. Compared with other ensemble methods, MEAT not only achieves the best robustness with negligible additional computational overhead but also minimizes the robust overfitting phenomenon. Finally, we conducted extensive experiments to demonstrate the effectiveness of MEAT.

\section{Acknowledgements}
\label{sec:acknowledgements}
This work was partly supported by the National Natural Science Foundation of China under Grant Nos. 62202104, 62102422, 62072109 and U1804263; the Ministry of Science and Technology, Taiwan, under Grant MOST 111-2628-E-155-003-MY3; and Youth Foundation of Fujian Province, P.R.China, under Grant No.2021J05129.



\vfill\pagebreak

\bibliographystyle{IEEEbib}
\bibliography{strings,refs}

\end{document}